\documentclass[letterpaper]{article} 
\usepackage{aaai2026}  
\usepackage{times}  
\usepackage{helvet}  
\usepackage{courier}  
\usepackage[hyphens]{url}  
\usepackage{graphicx} 
\usepackage{natbib}  
\usepackage{caption} 
\usepackage{algorithm}
\usepackage{algorithmic}

\usepackage{amsmath}
\usepackage{amssymb}
\usepackage{booktabs}
\usepackage{multirow}
\usepackage{tikz}
\newcommand{\circled}[1]{\tikz[baseline=(char.base)]{
  \node[shape=circle,draw,inner sep=1pt] (char) {#1};}}
\usepackage{newfloat}
\usepackage{listings}
\usepackage{times}
\usepackage{helvet}
\usepackage{courier}
\usepackage{xcolor}
\DeclareCaptionStyle{ruled}{labelfont=normalfont,labelsep=colon,strut=off} 
\floatstyle{ruled}
\newfloat{listing}{tb}{lst}{}
\floatname{listing}{Listing}
\title{Combining LLM Semantic Reasoning with GNN Structural Modeling for Multi-View Multi-Label Feature Selection}

\author{
    Zhiqi Chen\textsuperscript{\rm 1,2},
    Yuzhou Liu\textsuperscript{\rm 1,2}\thanks{Corresponding author},
	Jiarui Liu\textsuperscript{\rm 1,2},
    Wanfu Gao\textsuperscript{\rm 1,2}\footnotemark[1]
}
\affiliations{
    \textsuperscript{\rm 1}College of Computer Science and Technology, Jilin University, China\\
	\textsuperscript{\rm 2}Key Laboratory of Symbolic Computation and Knowledge Engineering of Ministry of Education, Jilin University, China\\
    chenzq24@mails.jlu.edu.cn, liuyuzhou@jlu.edu.cn, liujr24@mails.jlu.edu.cn, gaowf@jlu.edu.cn
}

\usepackage{bibentry}

\begin{document}

\maketitle

\begin{abstract}
Multi-view multi-label feature selection aims to identify informative features from heterogeneous views, where each sample is associated with multiple interdependent labels. This problem is particularly important in machine learning involving high-dimensional, multimodal data such as social media, bioinformatics or recommendation systems. Existing Multi-View Multi-Label Feature Selection (MVMLFS) methods mainly focus on analyzing statistical information of data, but seldom consider semantic information. In this paper, we aim to use these two types of information jointly and propose a method that combines Large Language Models (LLMs) semantic reasoning with Graph Neural Networks (GNNs) structural modeling for MVMLFS. Specifically, the method consists of three main components. (1) LLM is first used as an evaluation agent to assess the latent semantic relevance among feature, view, and label descriptions. (2) A semantic-aware heterogeneous graph with two levels is designed to represent relations among features, views, and labels: one is a semantic graph representing semantic relations, and the other is a statistical graph. (3) A lightweight Graph Attention Network (GAT) is applied to learn node embedding in the heterogeneous graph as feature saliency scores for ranking and selection. Experimental results on multiple benchmark datasets demonstrate the superiority of our method over state-of-the-art baselines, and it is still effective when applied to small-scale datasets, showcasing its robustness, flexibility, and generalization ability.

\end{abstract}


\begin{figure}[t]
\centering
\includegraphics[width=0.8\columnwidth]{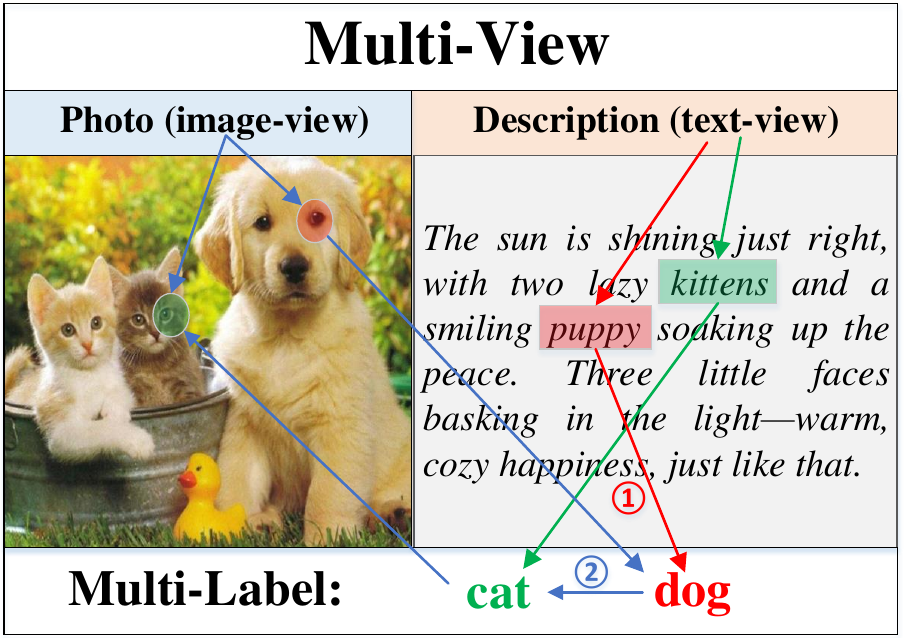} 
\caption{An example of semantic relations in MVMLFS. \circled{1} In text view, feature “\textit{puppy}” has a similar meaning to the label “\textit{dog}”, which can indicate it is an important feature for “\textit{dog}”. \circled{2} In image view, since labels “\textit{dog}” and “\textit{cat}” have high semantic similarity, feature “\textit{eye}” (which is informative for recognizing “\textit{dog}”) may also be informative for “\textit{cat}”.}
\label{fig1}
\end{figure}

\begin{figure*}[t]
\centering
\includegraphics[width=0.9\textwidth]{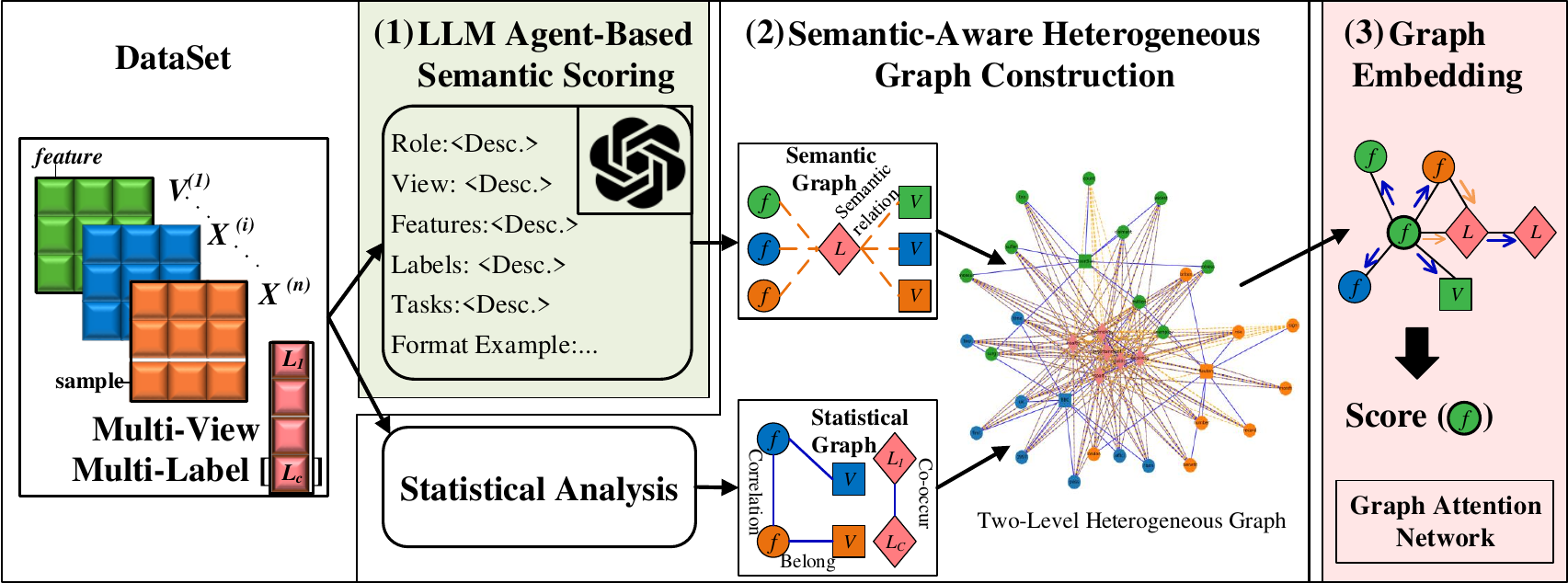} 
\caption{Framework of our method. It consists of three main steps: (1) establishing semantic relations among features, views, and labels by LLM; (2) combining semantic and statistical information, a two-level heterogeneous graph is constructed as the structural modeling of the dataset; (3) employing Graph Attention Network to learn graph embeddings, enabling the estimation of importance scores for feature nodes.
}
\label{fig2}
\end{figure*}

\section{Introduction}
With the rapid development of machine learning, people tend to associate multi-modal data sources (typically such as texts and images) to train more powerful models, so multi-view multi-label learning has attracted increasing attention in various domains such as social media, bioinformatics, or recommendation systems \cite{ref1,ref58,ref56}. However, the combination of different data sources not only gives useful information, but also introduces noisy, irrelevant, or redundant features \cite{ref36, ref43,ref27}. Thus, Multi-View Multi-Label Feature Selection (MVMLFS), aiming at identifying informative features from multi-modal data sources to reduce dimensionality, becomes critical for the learning process to shorten training time, enhance generalization, avoid overfitting, and finally improve performance \cite{ref18,ref28}.

To better complete the MVMLFS task, many methods have been given in recent years, such as MSFS~\cite{ref18}, M2LD \cite{ref40}, ELSMML \cite{ref41}, DHLI \cite{ref26}, and EF2FS \cite{ref25}. Although these methods achieve good performance, they usually rely on statistical methods, which means they focus on analyzing statistical information, such as the distributions of data \cite{ref29, ref46}. However, semantic information, as a kind of important information in many research fields, has seldom been considered in feature selection \cite{ref48,ref45,ref31}.

Our key insight is that semantic information can also be useful to construct extra relations for helping feature selection. As illustrated in Figure 1, on one hand, the semantic relations between features and labels can affect the value of features directly, such as: according to the similar meaning of words puppy and dog, the feature “puppy” is important for the label “dog”. On the other hand, the feature importance can be transmitted based on the semantic relations, for example, semantically similar labels such as “dog” and “cat” often have high semantic relevance and share common informative features, reflecting their underlying semantic correlation. Thus, since the feature “eye” is informative for the label “dog”, it may also be informative for the label “cat”. In this paper, we intend to introduce semantic information to the MVMLFS process. To achieve this goal, two main issues need to be solved.
\begin{itemize}
\item How to gain proper semantic information of features, views or labels? In some situations, these objects are described by natural language, and they need enough domain knowledge to understand their meanings. While in other situations, these objects (especially features) may be named by just sequence numbers, which limits the usability of semantic information.
\item How to effectively use the semantic information for supporting MVMLFS? On one hand, relying only on semantic information may not be enough, and it needs to be combined with other statistical information about the data. On the other hand, the high-dimensional features mean a high cost of semantic analysis, and it is necessary to balance the payback and cost.
\end{itemize}

To address these issues, we propose a method that combines Large Language Models (LLMs) semantic reasoning with Graph Neural Networks (GNNs) structural modeling for MVMLFS. Specifically, the method consists of three main components, as shown in Figure 2.

Firstly, semantic scoring evaluation based on LLMs. Benefiting from the great power of LLMs on semantic reasoning, we use them as the agent to assess the latent semantic relevance among feature, view, and label definitions. For the feature with a meaningless name, we use “view+number” to gain its “pseudo-semantic” information by considering that the view often has its clear semantics, and features in one view have the same semantic space.

Secondly, semantic-aware heterogeneous graph construction. GNNs have particular advantages in structural modeling for relation analysis. Thus, we design a two-level heterogeneous graph, in which one is semantic graph representing semantic relations among features, views, and labels; the other is statistical graph consisting of three kind of edges to represent different relations, including belong relations between features and views, correlations among features, views, and labels calculated by mutual information, and co-occurrence among labels \cite{ref47,ref57}. In this way, the heterogeneous graph is constructed to integrate the semantic and statistical information of data.

Finally, embedding-based feature importance estimation. A lightweight Graph Attention Network (GAT) is applied to learn node embedding as feature importance scores. The reasons for using GAT are: it can learn different correlation degrees from different types of edges, and it considers the different importance of various neighbors to the central node. According to the importance scores, it ranks and selects the most informative features across all views.

In summary, our main contributions are given as follows: 
\begin{itemize}
\item To the best of our knowledge, this is the first attempt to leverage LLMs to introduce extra semantic information for supporting multi-view multi-label feature selection. 
\item A semantic-aware two-level heterogeneous graph is constructed for supporting the joint analysis of new semantic information with traditional statistical information. Moreover, GAT is used to learn the node embedding in a heterogeneous graph for quantifying the importance of features, so that the selection can better consider the global information. 
\item The extensive experiments on benchmark multi-view multi-label datasets demonstrate the superiority of our method over state-of-the-art baselines. Moreover, it shows the effectiveness of LLMs and GNNs and can also achieve good performance on the small-scale datasets. 
\end{itemize}

\section{Related Work}

\subsection{Multi-View Multi-Label Feature Selection}
Multi-label feature selection has been extensively studied to reduce input dimensionality while preserving label dependency structures \cite{ref23, ref32,ref33}. In multi-view scenarios, the challenge intensifies due to feature heterogeneity and view complementarity \cite{ref38, ref39, ref42}. View-independent methods treat each view separately and fuse results post hoc, ignoring cross-view label interactions \cite{ref2, ref34, ref37}. View-concatenation methods concatenate all views into a single matrix before selection \cite{ref3,ref21}, which may introduce noise and redundancy. Joint optimization frameworks learn feature relevance jointly across views and labels using sparse regularization \cite{ref4}, matrix factorization \cite{ref5}, or tensor-based formulations \cite{ref6}. However, most of the methods are limited by linear assumptions and lack semantic alignment across views and labels.

\subsection{Graph-Based Feature Selection}
Graphs have emerged powerfully for modeling relational structures among features, samples, and labels. Early works constructed feature similarity graphs \cite{ref7}, while recent advances explore label-aware or semantic graphs \cite{ref8, ref44}. Graph convolutional networks and graph attention networks further enhance this line by enabling task-driven embedding learning \cite{ref9,ref34,ref35}. Some studies leverage label co-occurrence graphs \cite{ref10} or sample-feature graphs \cite{ref11} to enhance selection, yet they often overlook cross-modal or cross-view semantics in feature-label relations. Our work extends this by introducing multi-type heterogeneous graphs with semantic edges and LLM guidance.

\subsection{Large Language Models for Semantic Scoring}
With the rapid development of natural language processing, LLMs (e.g., GPT, LLaMA) have shown strong potential in aligning text representations and enabling semantic inference \cite{ref13, ref49, ref50}. Recent research has utilized LLMs for prompt-driven data labeling \cite{ref14}, knowledge graph construction \cite{ref15}, feature recommendation \cite{ref16}, and few-shot tabular learning \cite{ref30}. However, LLMs are seldom used for fine-grained semantic edge construction in feature-label graphs, especially in multi-view settings.
Unlike prior work that treats LLMs as end-to-end black-box classifiers, our method leverages LLMs as semantic scoring agents, refining features, labels, and views connections via natural language prompts and integrating them into downstream graph learning.

While existing works contribute to either multi-view feature selection, graph modeling, or LLM-guided tasks, the proposed method uniquely combines all three. A semantic-aware heterogeneous graph is constructed with typed nodes and edges, integrating LLM-derived semantic priors with GAT-based structural learning. This novel integration enables a principled yet practical method for multi-view multi-label feature selection in semantically complex domains.

\section{Methodology}

\subsection{Problem Formulation}

Since our method introduces extra semantic information into MVMLFS, we reformulate the problem by incorporating natural language descriptions of features, views, and labels.

Given a multi-view multi-label dataset $D = \left\{ F^{(i)} \in \mathbb{R}^{n \times d{(i)}} \right\}_{i=1}^V$, with $V$ different views, and a binary label matrix is denoted by ${L} \in \{0,1\}^{n \times d{(L)}}$, where $n$ is the number of instances, $d{(L)}$ is the number of labels, and  $F^{(i)}$ is the $i$-th view with $d(i)$-dimensional features.
To simplify the description, we denote the set of features in the $i$-th view as $F^{(i)}$, then $f^{(i)}_m \in F^{(i)}$ denotes the $m$-th feature in the $i$-th view, and all the features in the dataset is denoted as $F = \bigcup_{i=1}^{V} F^{(i)}$. Similarly,  $l_j \in {L}$ denotes the $j$-th label.

Moreover, to enhance semantic understanding, we associate each feature, view, and label with a corresponding textual description in natural language, we have three kinds of text sets: (1) $T^F = \left\{ T_i^f \right\}_{i=1}^V$, where $T_i^f = \{ t_1, t_2, \dots, t_{d{(i)}} \}$ is a set of texts describing the $d{(i)}$ features in the $i$-th view. (2) $T^V = \{ t_1, t_2, \dots, t_V \}$ gives text descriptions of the $V$ views. (3) $T^L = \{ t_1, t_2, \dots, t_{d{(L)}}\}$ gives texts describing the meaning of $d(L)$ labels.

Based on the above definitions, the objective is to select a subset of discriminative features from each view and finally aggregate them into a global set of features.

\subsection{Semantic Score Computation Based on LLM}

To incorporate semantic understanding into the feature selection process, we leverage an LLM to quantify semantic scores that capture the relevance among features, views, and labels. This semantic scoring mechanism complements statistical structures by introducing language-level conceptual associations, enabling more informed selection decisions across heterogeneous views.

To achieve the goal, an LLM-based agent is designed to perform semantic analysis on the description texts of features, views, and labels. In this process, we give structured prompts tailored for semantic scoring. Although the detailed contents are adjusted for different analyzing objects (such as feature-label, view-label, or label-label), the prompt mainly consists of five key components, that is $P =\{Role + (Views, Features, Labels) + Task\}$: 
\begin{itemize}
\item $Role$, defines the thinking perspective of LLM, \textit{“You are a data scientist working on multi-view multi-label feature selection. Your goal is dedicated to exploring the impact of semantic relationships among features, labels, and views on feature selection results.”}
\item $Views$, the names and the description texts of the views.
\item $Features$, the list of their names and descriptions. Note that if the feature name is without explicit semantics, the pseudo-semantic “view+number” is adopted instead.
\item $Labels$, the list of label names representing semantic categories.
\item $Tasks$, it needs to identify semantically relevant or redundant features and assign semantic scores in the range [0, 1] between the given objects.
\end{itemize}

According to the prompt, when given any two objects (feature, view, or label), the LLM-agent returns a semantic score between them, denoted by $LLMScore$. For example, given a feature $f_m^{(i)}$ and a label $l_j$, semantic score is given as: 
\[
LLMScore(f^{(i)}_m, l_j) = 
\]
\[
P(Role + (t_i \in T^v, t_m \in T^f_i, t_j \in T^L) + Task)
\]

In this way, we can gain semantic scores to establish extra relations among features, views, and labels, providing additional guidance for the subsequent feature selection process.

\subsection{Semantic-Aware Heterogeneous Graph Construction}
To comprehensively capture both data-driven and knowledge-driven relationships, we construct a unified heterogeneous graph $\mathcal{G}=(\mathcal{V}, \mathcal{E})$ by integrating statistical dependencies and semantic correlations. This graph comprises three types of nodes \emph{views}, \emph{features}, and \emph{labels} as well as multiple types of edges, each reflecting a distinct relational perspective.

\subsubsection{Semantic Graph Construction}

Beyond capturing statistical dependencies, we construct a semantic graph to model implicit semantic associations among features, labels, and views. This graph complements the statistical structure by introducing edges derived from a language model-based semantic scoring mechanism. Three types of semantic relations are considered: \textit{feature-label}, \textit{view-label}, and \textit{label-label} semantic edges. Edges are established between nodes with scores exceeding a predefined threshold $\delta$, and their weights are set proportionally to their semantic relevance. For example, the semantic feature-label edge is defined as:
\[
E_{{fls}} = \{(f_m^{(i)}, l_j) \mid LLMScore^{(i)}_{mj} > \delta\}
\]

Similarly, semantic edges between views and labels, and between labels themselves, are generated in the same manner. These semantic relations are fused into the heterogeneous graph structure. And unified semantic graph captures multi-level dependencies among features, views, and labels, providing enriched structural signals for guiding downstream feature selection.

\subsubsection{Statistical Graph Construction}

To model the statistical relationships among views, features, and labels, we construct a graph $\mathcal{G}_s$, where edges encode different types of statistical dependencies. Specifically, we define the following edge types and their associated weights:

\paragraph{(1) Feature-View Belonging Edges.}
Each feature $f_i$ is inherently associated with a specific view $v$. We connect them with an unweighted edge:
\[
E_{fv} = \{(f_i, v)\ |\ f_i \in F_v\}
\]

\paragraph{(2) Feature-Label Correlation Edges.}
We compute the mutual information (MI) between each feature $f_i \in F_v$ and label $l_j \in {L}$. An edge is added between $f_i$ and $l_j$ if ${MI}(f_i, l_j) > \tau_1$. The edge weight is defined as the normalized MI value:
{\small
\[
E_{fl} = \{(f_i, l_j)\ |\ {MI}(f_i, l_j) > \tau_1\}, w_{fl} = \frac{{MI}(f_i, l_j)}{\max_{i,j}{MI}(f_i, l_j)}
\]
}
\paragraph{(3) Feature-Feature Correlation Edges.}
To capture relevance among features, we calculate MI for each feature pair $(f_i, f_j)$. An edge is added if ${MI}(f_i, f_j) > \tau_2$. The edge weight reflects the normalized correlation strength:
{\small
\[
E_{ff} = \{(f_i, f_j)\ | {MI}(f_i, f_j) > \tau_2\},  w_{ff} = \frac{{MI}(f_i, f_j)}{\max_{i,j}{MI}(f_i, f_j)}
\]
}

\paragraph{(4) Label-Label Co-occurrence Edges.}
We compute co-occurrence statistics across the label matrix $L$. For labels $l_i$ and $l_j$, let $c_{ij} = \sum_{n=1}^{N} \mathbb{I}[L_{ni} = 1 \land L_{nj} = 1]$. An edge is added if $c_{ij} > 0$, with weight normalized across all label pairs:
\[
E_{ll} = \{(l_i, l_j)\ |\ c_{ij} > 0\}, \quad w_{ll} = \frac{c_{ij}}{\max_{i,j} c_{ij}}
\]

\paragraph{(5) View-Label Correlation Edges.}
To capture the overall relationship between views and labels, we compute the average mutual information between all features in a view $v$ and label $l_j$:
\[
\bar{M}_{vj} = \frac{1}{|{F}_v|} \sum_{f_i \in {F}_v}{MI}(f_i, l_j)
\]

If $\bar{M}_{vj} > \tau_3$, we add an edge, with the weight defined as:
\[
E_{vl} = \{(v, l_j)\ |\ \bar{M}_{vj} > \tau_3\}, \quad w_{vl} = \frac{\bar{M}_{vj}}{\max_{v,j} \bar{M}_{vj}}
\]

Through this construction, a weighted statistical graph is obtained that captures rich multi-type dependencies across views, features, and labels. These weighted edges serve as informative priors for subsequent feature selection.

\subsection{Type-Aware Heterogeneous Graph Attention Network}
To effectively model the intricate semantic interactions and deliver messages among views, features, and labels, a type-aware Heterogeneous Graph Attention Network is adopted that extends the standard GAT to heterogeneous graphs. This is achieved via a relation-specific convolution framework that captures type-aware message passing across distinct node and edge types.

Specifically, for each edge relation $r = (\tau_s, \tau_r, \tau_t)$, which represents a connection from a source node of type $\tau_s$ to a target node of type $\tau_t$ through relation type $\tau_r$, a dedicated GAT layer is instantiated. The embedding of a target node $v$ of type $\tau_t$ is computed as:
\[
H_v^{(\tau_t)} = \sum_{(\tau_s, \tau_r, \tau_t)} {GAT}^{(\tau_r)}(h_u^{(\tau_s)}, h_v^{(\tau_t)})
\]

Where $h_u^{(\tau_s)}$ and $h_v^{(\tau_t)}$ denote the input features of neighboring source and target nodes, and $\text{GAT}^{(\tau_r)}$ is a relation-specific attention mechanism. For computational efficiency, we adopt single-head attention and omit self-loops.

The outputs from different relation types targeting the same node type are aggregated via summation:
\[
H^{(\tau)} = \sum_{r \in \mathcal{R}_{\rightarrow \tau}} H^{(\tau)}_r
\]

Where $\mathcal{R}_{\rightarrow \tau}$ is the set of all relations pointing to node type $\tau$.

Once node representations are computed, we derive feature importance scores from the learned embeddings of feature nodes. For each feature node $f$, its relevance score is defined as:
\[
{score}(f) = h_f
\]

Where $h_f \in \mathbb{R}$ is the scalar output embedding of node $f$.

We rank all feature nodes based on their scores and select the top-$k$ features to form the final feature subset:

\[
\mathcal{S}_k = {TopK}\left(\{{score}(f) \mid f \in \mathcal{V}_{{feature}}\},\ k\right)
\]

\begin{table*}[t] 
\small
\centering
\label{tab:datasets}
\begin{tabular}{lccccccc}
\toprule
\textbf{Feature views} & \textbf{SCENE} & \textbf{VOC07} & \textbf{MIRFlickr} & \textbf{OBJECT} & \textbf{3Sources}& \textbf{Yeast} \\
\midrule
View1($d_1$)          & CH(64)         & DH(100)        & DH(100)        & CH(64)         & BBC(1000)& GE(79) \\
View2($d_2$)          & CM(225)        & GIST(512)      & GIST(512)      & CM(225)        & Reuters(1000)& PP(24)\\
View3($d_3$)          & CORR(144)      & HH(100)        & HH(100)        & CORR(144)   & Guardian(1000)   & --   \\
View4($d_4$)          & EDH(73)        & --             & --             & EDH(73)        & --    & --\\
View5($d_5$)          & WT(128)        & --             & --             & WT(128)        & --    & --\\
\midrule
Samples($n$)          & 4400           & 3817           & 4053             & 6047           & 169& 2417 \\
Features($d$)         & 634            & 712            & 712              & 634            & 3000& 103\\
Labels($c$)           & 33             & 20             & 38               & 31             & 6  & 14\\
\bottomrule
\end{tabular}

\caption{Characteristics of the datasets in our experiments.}
\end{table*}

\section{Experiments}

\subsection{Experimental Setup}
\paragraph{Datasets} 

We evaluate the proposed method on six publicly available multi-view multi-label datasets: SCENE \cite{ref52}, VOC07 \cite{ref53}, MIRFlickr \cite{ref54}, OBJECT \cite{ref52}, 3Sources \cite{ref55}, and Yeast \cite{ref51}, which cover various domains including vision, news, and genes with different available semantic information. The key characteristics of these datasets are summarized in Table~1. Each view is identified by its corresponding individual name, with the dimensionality shown in braces. Missing views are denoted by `--'. The number of samples, features, and labels is denoted by \( n \), \( d \), and \( c \), respectively.

\paragraph{Comparing Methods} 
We compare our method with three representative multi-label feature selection methods based on information theory, including ENM \cite{ref20}, MLSMFS \cite{ref22}, and STFS \cite{ref19}. In addition, we evaluate three multi-view multi-label feature selection methods, including MSFS \cite{ref18}, DHLI \cite{ref26}, and EF2FS \cite{ref25}. Notably, the adopted learning framework also enables the extraction of a feature importance matrix, which provides insights into the relative contribution of each feature to the final prediction.

\paragraph{Evaluation Metrics} 

For all datasets, MLKNN with \( k = 10 \) is employed as the base classifier. Feature selection is performed with selection ratios ranging from 2\% to 20\% of the total number of features, increasing in 2\% increments. In each experiment, 70\% of the samples are randomly selected for training and the remaining 30\% for testing. To ensure robustness, each experiment is repeated 10 times, and the mean performance along with the standard deviation is reported. To evaluate the effectiveness of feature selection, four widely used multi-label classification metrics are adopted~\cite{ref23, ref24}: Average Precision (AP), Macro-Area Under the Curve (AUC), Label Ranking Average Precision (LRAP), and Hamming Loss (HL). For AP, AUC, and LRAP, higher values indicate better performance, whereas lower HL values reflect better outcomes. These metrics collectively assess how well the selected features preserve the predictive power with respect to the label space.

\begin{table*}[t]  
\centering
\small
\renewcommand{\arraystretch}{1.25}
\setlength{\tabcolsep}{1mm}{
\begin{tabular}{l|ccccccc}
\toprule
\textbf{Datasets}  &\textbf{Ours} & \textbf{ENM} & \textbf{MLSMFS} & \textbf{STFS} & \textbf{MSFS} & \textbf{DHLI} & \textbf{EF2FS}\\
\hline
\multicolumn{8}{c}{\textit{AP} $\uparrow$} \\
\hline
SCENE      & \textbf{0.2888$\pm$0.016} & 0.2274$\pm$0.014 & 0.2308$\pm$0.013 & 0.2290$\pm$0.014 & 0.2328$\pm$0.010 & 0.2440$\pm$0.016 & \underline{0.2499$\pm$0.006} \\
\hline
VOC07      & \textbf{0.1801$\pm$0.021} & 0.1314$\pm$0.005 & 0.1155$\pm$0.002 & \underline{0.1358$\pm$0.007} & 0.1114$\pm$0.001 & 0.1280$\pm$0.002 & 0.1135$\pm$0.003 \\
\hline
MIRFlickr  & \textbf{0.3131$\pm$0.027} & \underline{0.2931$\pm$0.011} & 0.2599$\pm$0.001 & 0.2879$\pm$0.012 & 0.2434$\pm$0.001 & 0.2545$\pm$0.003 & 0.2692$\pm$0.004 \\
\hline
OBJECT     & \textbf{0.1746$\pm$0.013} & 0.1022$\pm$0.013 & 0.0981$\pm$0.012 & 0.1040$\pm$0.011 & 0.1067$\pm$0.010 & 0.1174$\pm$0.019 & \underline{0.1252$\pm$0.010} \\
\hline
3Sources   & \textbf{0.2086$\pm$0.021} & \underline{0.2044$\pm$0.003} & 0.1880$\pm$0.009 & 0.1946$\pm$0.014 & 0.1925$\pm$0.001 & 0.2033$\pm$0.004 & 0.1928$\pm$0.002 \\
\hline
Yeast      & \textbf{0.3398$\pm$0.025} & \underline{0.3231$\pm$0.010} & 0.3164$\pm$0.011 & 0.3141$\pm$0.012 & 0.3109$\pm$0.003 & 0.3215$\pm$0.006 & 0.3072$\pm$0.007 \\										
\hline
\multicolumn{8}{c}{\textit{AUC} $\uparrow$} \\
\hline
SCENE & \textbf{0.6581$\pm$0.021} & 0.6026$\pm$0.033 & 0.5965$\pm$0.024 & 0.6125$\pm$0.037 & 0.5929$\pm$0.023 & \underline{0.6215$\pm$0.032} & 0.6070$\pm$0.014 \\
\hline
VOC07 & \textbf{0.6410$\pm$0.033} & 0.5947$\pm$0.020 & 0.5363$\pm$0.006 & \underline{0.6087$\pm$0.021} & 0.5001$\pm$0.001 & 0.5754$\pm$0.008 & 0.5066$\pm$0.008 \\
\hline
MIRFlickr & \textbf{0.6151$\pm$0.034} & \underline{0.6132$\pm$0.020} & 0.5501$\pm$0.001 & 0.6029$\pm$0.015 & 0.5091$\pm$0.001 & 0.5457$\pm$0.005 & 0.5642$\pm$0.009 \\
\hline
OBJECT & \textbf{0.6557$\pm$0.016} & 0.6193$\pm$0.030 & 0.6108$\pm$0.023 & 0.6188$\pm$0.028 & 0.5947$\pm$0.019 & \underline{0.6339$\pm$0.033} & 0.6279$\pm$0.011  \\
\hline
3Sources & \textbf{0.5116$\pm$0.041} & \underline{0.5036$\pm$0.011} & 0.4828$\pm$0.026 & 0.5028$\pm$0.011 & 0.5004$\pm$0.002 & 0.4963$\pm$0.011 & 0.5004$\pm$0.003 \\	
\hline
Yeast & \textbf{0.5574$\pm$0.029} & 0.5395$\pm$0.021 & \underline{0.5422$\pm$0.018} & 0.5365$\pm$0.020 & 0.5118$\pm$0.007 & 0.5364$\pm$0.016 & 0.5248$\pm$0.010 \\								
\bottomrule
\end{tabular}
}
\caption{Experimental results (mean ± std) for AP and AUC, where the 1st/2nd best results are
shown in boldface/underline.}
\end{table*}

\begin{table*}[t] 
\centering
\small
\renewcommand{\arraystretch}{1.25}
\setlength{\tabcolsep}{1mm}{
\begin{tabular}{l|ccccccc}
\toprule
\textbf{Datasets} & \textbf{Ours} & \textbf{ENM} & \textbf{MLSMFS} & \textbf{STFS} & \textbf{MSFS} & \textbf{DHLI} & \textbf{EF2FS}\\
\hline
\multicolumn{8}{c}{\textit{LRAP} $\uparrow$} \\
\hline
SCENE & \underline{0.7930$\pm$0.009} & 0.7848$\pm$0.012 & 0.7810$\pm$0.008 & 0.7834$\pm$0.013 & 0.7763$\pm$0.012 & 0.7890$\pm$0.011 & \textbf{0.8087$\pm$0.007} \\
\hline
VOC07 & \textbf{0.5944$\pm$0.017} & 0.5883$\pm$0.008 & 0.5584$\pm$0.002 & \underline{0.5929$\pm$0.011} & 0.5538$\pm$0.014 & 0.5834$\pm$0.003 & 0.5537$\pm$0.019 \\
\hline
MIRFlickr & \underline{0.6795$\pm$0.024} & \textbf{0.6877$\pm$0.010} & 0.6514$\pm$0.001 & 0.6760$\pm$0.012 & 0.5704$\pm$0.053 & 0.6556$\pm$0.005 & 0.6513$\pm$0.013 \\
\hline
OBJECT & \textbf{0.4910$\pm$0.019} & 0.4287$\pm$0.026 & 0.4088$\pm$0.011 & 0.4288$\pm$0.022 & 0.4098$\pm$0.018 & 0.4246$\pm$0.031 & \underline{0.4395$\pm$0.012} \\
\hline
3Sources & \textbf{0.4636$\pm$0.031} & \underline{0.4557$\pm$0.010} & 0.4275$\pm$0.050 & 0.4528$\pm$0.037 & 0.4041$\pm$0.023 & 0.4524$\pm$0.011 & 0.4442$\pm$0.017 \\
\hline
Yeast & 0.6583$\pm$0.020 & \textbf{0.6694$\pm$0.013} & 0.6638$\pm$0.014 & 0.6604$\pm$0.016 & 0.6520$\pm$0.025 & 0.6613$\pm$0.016 & \underline{0.6679$\pm$0.007} \\
\hline
\multicolumn{8}{c}{\textit{HL} $\downarrow$} \\
\hline
SCENE & \textbf{0.1019$\pm$0.003} & 0.1060$\pm$0.007 & 0.1070$\pm$0.005 & 0.1050$\pm$0.007 & 0.1104$\pm$0.004 & \underline{0.1025$\pm$0.005} & 0.1074$\pm$0.004 \\
\hline
VOC07 & \textbf{0.0831$\pm$0.002} & 0.0850$\pm$0.001 & 0.0862$\pm$0.001 & 0.0859$\pm$0.002 & \underline{0.0847$\pm$0.002} & 0.0852$\pm$0.000 & 0.0963$\pm$0.014 \\
\hline
MIRFlickr & 0.1799$\pm$0.009 & \textbf{0.1710$\pm$0.005} & 0.1919$\pm$0.001 & \underline{0.1756$\pm$0.006} & 0.2288$\pm$0.022 & 0.1896$\pm$0.003 & 0.1934$\pm$0.004 \\
\hline
OBJECT & \underline{0.0587$\pm$0.001} & 0.0611$\pm$0.003 & 0.0590$\pm$0.002 & 0.0616$\pm$0.003 & 0.0654$\pm$0.003 & \textbf{0.0575$\pm$0.003} & 0.0594$\pm$0.002 \\
\hline
3Sources & \textbf{0.1850$\pm$0.009} & 0.2119$\pm$0.007 & 0.2647$\pm$0.043 & 0.2033$\pm$0.010 & 0.2112$\pm$0.017 & 0.2199$\pm$0.007 & \underline{0.1954$\pm$0.006} \\
\hline
Yeast & 0.2378$\pm$0.007 & \underline{0.2264$\pm$0.001} & \textbf{0.2247$\pm$0.003} & 0.2277$\pm$0.003 & 0.2607$\pm$0.006 & 0.2282$\pm$0.003 & 0.2436$\pm$0.008 \\
\bottomrule
\end{tabular}
}
\caption{Experimental results (mean ± std) for LRAP and HL, where the 1st/2nd best results are shown in boldface/underline.}
\end{table*}

\begin{figure}[t]
\centering
\includegraphics[width=1\columnwidth]{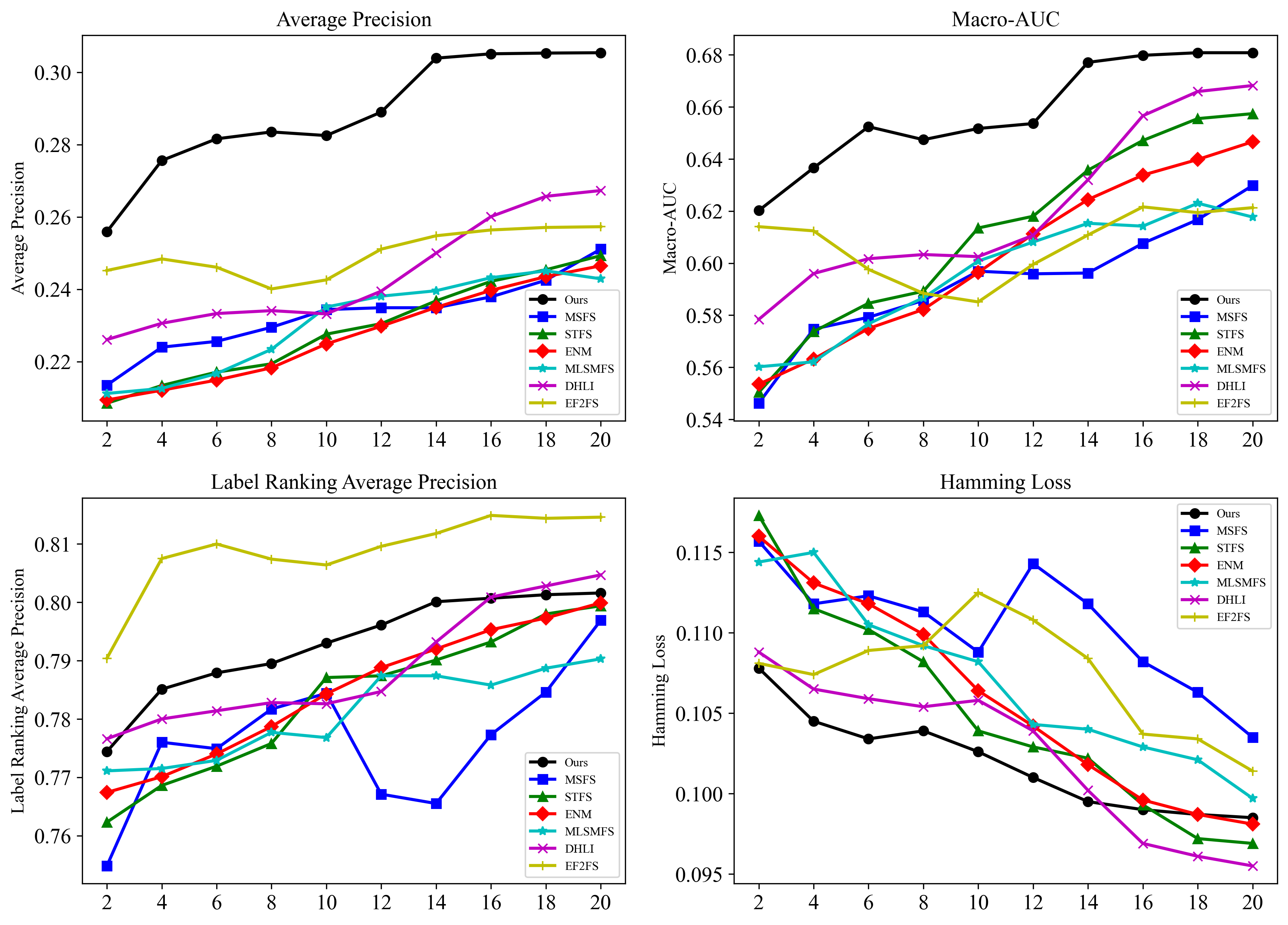} 
\caption{Seven methods on SCENE in terms of Average Precision, Macro-AUC, Label Ranking Average Precision, and Hamming loss.}
\label{fig3}
\end{figure}

\begin{figure}[t]
\centering
\includegraphics[width=1\columnwidth]{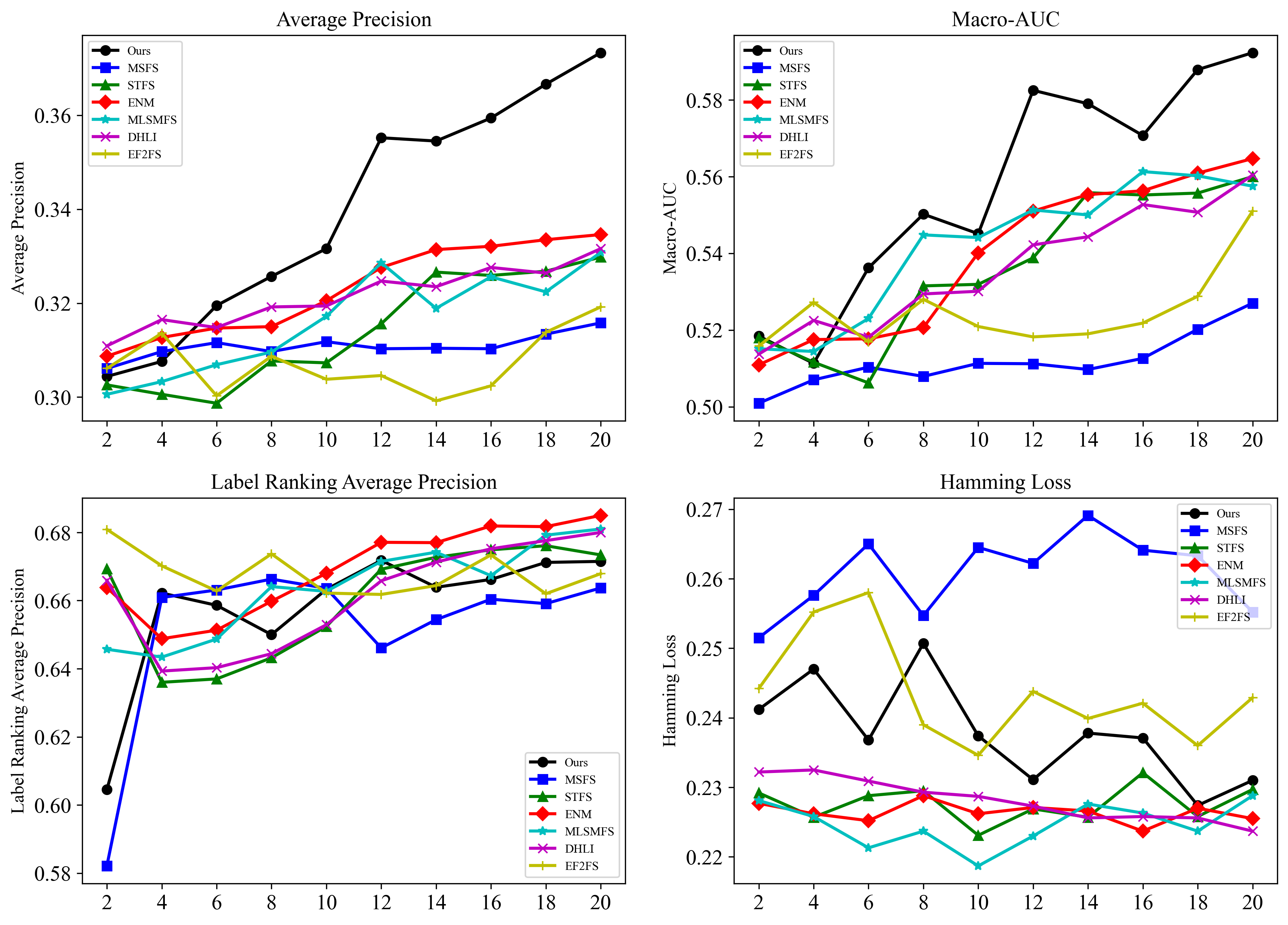} 
\caption{Seven methods on Yeast in terms of Average Precision, Macro-AUC, Label Ranking Average Precision, and Hamming loss.}
\label{fig4}
\end{figure}

\subsection{Results and Discussion}

Table 2 and Table 3 present the performance of our method implemented with GPT4omini, compared with other representative methods. As can be seen, our method achieves most of the superior performance compared to other baseline methods on six datasets under the evaluation metrics AP and AUC. Figure 3 and Figure 4 clearly illustrate our performance metrics on SCENE and Yeast datasets, respectively. 

Our method consistently achieves the best performance in terms of AP and AUC across all datasets. On SCENE, VOC07, MIRFlickr, OBJECT, 3Sources, and Yeast, it outperforms the second-best method in AP by margins of +3.9\%, +4.4\%, +2.0\%, +4.9\%, +0.4\%, and +1.7\%, respectively, highlighting its ability to capture discriminative features in diverse, high-dimensional multi-label data. Regarding AUC, our method achieves 0.6410 on VOC07 and 0.6151 on MIRFlickr, demonstrating strong robustness in ranking relevant labels. On the OBJECT dataset, it obtains an AUC of 0.6557, surpassing the closest competitor by more than 2\%. In terms of LRAP, our method consistently ranks among the top performers on VOC07, OBJECT, and 3Sources, and second on SCENE and MIRFlickr, indicating enhanced feature selection in complex multi-view settings. For HL, our method attains the lowest values on SCENE, VOC07, and 3Sources, and ranks second on OBJECT. 

On Yeast dataset, which involves more complex label semantics, our method does not achieve the best performance in LRAP and HL, with differences of 1.11\% and 0.0131 from the top results, respectively. This can be attributed to the challenges in semantic perception posed by intricate label semantics, which limit further gains from the semantic-guided feature selection. As shown in Table 4, the comparable performance of the statistics-based variant without semantic enhancement further supports this observation. Overall, the results demonstrate the consistent superiority of our method over existing baselines.

\subsection{Ablation Study}
To investigate the effectiveness of different components in our method and the impact of different abilities of LLMs, we conduct an ablation study across six datasets, the experiments are shown in Table 4 and Figure 5.

Firstly, the impact of generating semantic scores of different LLMs is evaluated. Two popular models of different scales, GPT4omini and Deepseek R1-14B, are selected for our experiments. Among the variants, the model equipped with GPT4omini achieves the best overall performance, demonstrating that lightweight models can still provide reliable semantic understanding for feature selection.

Secondly, we assess the contribution of statistical structure by removing the semantic component. This configuration consistently underperforms compared to the full model, especially on the 3Sources dataset, which is a text-rich dataset where semantics play a crucial role. This highlights the importance of integrating semantic information, particularly in language-centric tasks.

\begin{figure}[t]
\centering
\includegraphics[width=1\columnwidth]{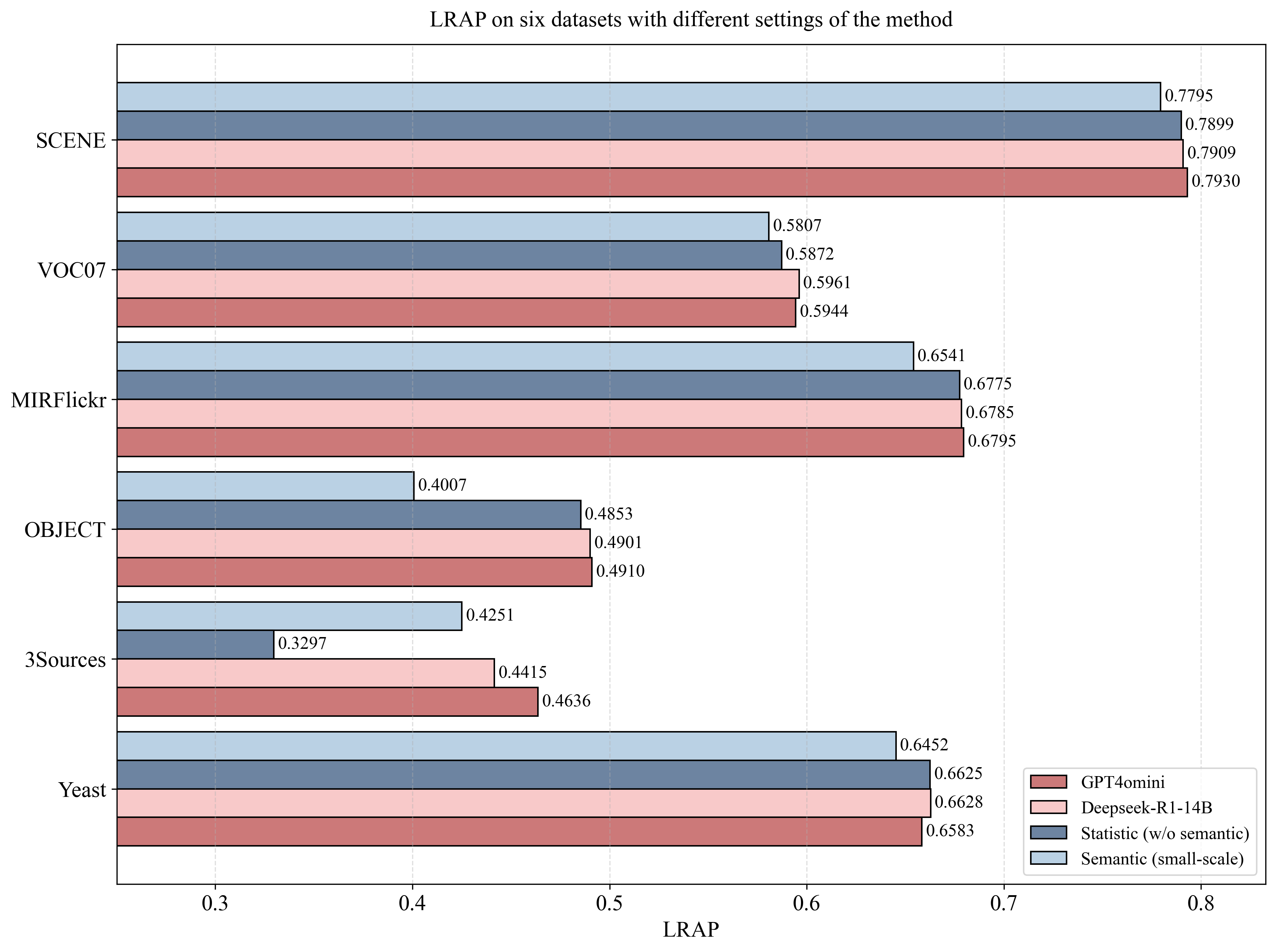} 
\caption{Ablation experimental results in terms of LRAP on six datasets with different settings of the method.}
\label{fig4}
\end{figure}
\begin{table}[t]
\centering
\small
\renewcommand{\arraystretch}{1.25}
\setlength{\tabcolsep}{0.3mm}{
\begin{tabular}{l|cccccc}
\toprule
Datasets & Metric & GPT & Deepseek & Statistical & Semantic \\
&&4omini&R1-14B&(w/o semantic)&(small-scale)\\
\hline
\multirow{4}{*}{\centering SCENE}    
  & AP& 	\textbf{0.2888}& 	\underline{0.2875}& 0.2842& 0.2310 \\
& AUC&  	\textbf{0.6581}& 	0.6512& 	\underline{0.6523}& 	0.6048 \\
 &LRAP&  \textbf{0.7930} & \underline{0.7909} & 0.7899 & 0.7795 \\
& HL& 	\textbf{0.1019}& 	\underline{0.1030}& 	0.1033& 	0.1059\\
\hline
\multirow{4}{*}{\centering Yeast}   
& AP& 	\underline{0.3398}& 	\textbf{0.3425}& 	0.3025& 	0.3001\\ 
& AUC& 	0.5574& 	\underline{0.5598}& 	\textbf{0.5628}& 	0.5042\\ 
&LRAP  & 0.6583 & \textbf{0.6628} & \underline{0.6625} & 0.6452 \\
& HL&  	0.2378& 	0.2353& 	\underline{0.2335}& 	\textbf{0.2334}\\ 
\bottomrule
\end{tabular}
}
\caption{Ablation experimental results on SCENE and Yeast.}
\end{table}

Finally, to evaluate the performance of the proposed method on small-scale datasets, we investigate whether semantic information is sufficient for effective feature selection. However, relying solely on semantics without incorporating statistical structure results in failed selection due to limited data support. To address this, we retain the semantic component with GPT4omini while introducing partial statistical information. Specifically, when constructing the statistical graph, only 50\% of the available data is used, simulating a small-scale dataset scenario. Interestingly, this variant achieves competitive results on small-scale dataset scenarios SCENE and Yeast, suggesting that semantic guidance is effective in small-scale dataset scenarios where statistical estimation is less reliable. These findings confirm the complementary benefits of combining statistical and semantic information in our feature selection method.

\section{Conclusion}
We propose a novel method for multi-view multi-label feature selection that integrates large language model priors with graph attention mechanisms. By constructing a heterogeneous graph that encodes both LLM-derived semantic information and statistical information based on mutual information as well as label co-occurrence, and applying a lightweight Graph Attention Network, the method produces informative feature embeddings for effective ranking and selection. Extensive experiments on benchmark datasets show consistent improvements over state-of-the-art methods, demonstrating superior performance, robustness, and generalization even on small-scale datasets. In future work, we will explore end-to-end optimization with LLM feedback for closed-loop refinement, integration of learned embeddings into downstream tasks, and self-supervised pretraining on heterogeneous graphs to enhance representation quality.

\section*{Acknowledgments}
The work is funded by Jilin Provincial Science and Technology Development Plan Project No.20240302084GX, and Changchun Science and Technology Bureau Project 23YQ05.

\bibliography{aaai2026}

\end{document}